\documentclass[11pt]{article}

\usepackage{amssymb}
\usepackage{latexsym}
\usepackage[mathscr]{euscript}
\usepackage{graphicx}
\usepackage{float}
\usepackage{subfigure}

\usepackage{amscd}
\usepackage{amsmath}
\usepackage{amssymb}
\usepackage{fullpage}
\usepackage{xcolor}

\usepackage{amssymb}
\usepackage{latexsym}
\usepackage[mathscr]{euscript}
\usepackage{pb-diagram}
\usepackage{latexsym}
\usepackage{amsfonts}
\usepackage{epsfig}

\usepackage{tikz}

\newcommand\bb[1]{\Bbb{#1}}

\newcommand\pr{^{\prime}}

\newcommand\bbz{\Bbb{Z}}

\makeatletter
\def\bbordermatrix#1{\begingroup \m@th
  \@tempdima 4.75\p@
  \setbox\z@\vbox{%
    \def\cr{\crcr\noalign{\kern2\p@\global\let\cr\endline}}%
    \ialign{$##$\hfil\kern2\p@\kern\@tempdima&\thinspace\hfil$##$\hfil
      &&\quad\hfil$##$\hfil\crcr
      \omit\strut\hfil\crcr\noalign{\kern-\baselineskip}%
      #1\crcr\omit\strut\cr}}%
  \setbox\tw@\vbox{\unvcopy\z@\global\setbox\@ne\lastbox}%
  \setbox\tw@\hbox{\unhbox\@ne\unskip\global\setbox\@ne\lastbox}%
  \setbox\tw@\hbox{$\kern\wd\@ne\kern-\@tempdima\left[\kern-\wd\@ne
    \global\setbox\@ne\vbox{\box\@ne\kern2\p@}%
    \vcenter{\kern-\ht\@ne\unvbox\z@\kern-\baselineskip}\,\right]$}%
  \null\;\vbox{\kern\ht\@ne\box\tw@}\endgroup}
\makeatother

\newcommand\commentout[1]{\marginpar{\tiny $\backslash$commentout}}

\newcommand\qed{\hfill$\square$}

\def\compcirc {\mbox{\hspace{.05cm}}\raisebox{.04cm}{\tiny  {$\circ$ }}}

\newtheorem{Definition}{Definition}[section]

\newtheorem{Example}{Example}[section]
\newtheorem{Remark}{Remark}[section]

\usepackage{graphics}
\title{Topological Approaches to Deep Learning}
\author{Gunnar Carlsson \\ Department of Mathematics, Stanford University\\Stanford, California 94305   \and Rickard Br\"{u}el Gabrielsson \\Department of Computer Science, Stanford University \\Stanford, California, 94305}
\begin{document}
\maketitle
\section{Introduction}  Deep neural networks \cite{priddy}  are a powerful and fascinating methodology for solving problems with large and complex data sets.  They use directed graphs as a template for very large computations, and have   demonstrated a great deal of success in the study of various kinds of data, including images, text, time series, and many others.  One issue that restricts their applicability, however, is the fact that it is not understood in any kind of detail how they work. A related problem is that there is often a certain kind of overfitting to particular data sets, which results in the possibility of so-called adversarial behavior, where they can be made to fail by making very small changes to image data that is almost imperceptible to a human.  For these reasons, it is very desirable to develop methods for gaining  understanding of the internal states of the neural networks. Because of the very large number of nodes (or neurons), and because of the stochastic nature of the optimization algorithms used to train  the networks, this is  a problem in data analysis, specifically for unsupervised data analysis.  The initial goal of the work in this paper was to perform topological data analysis (TDA) on the internal states of the neural nets being trained on image data to demonstrate that TDA can provide this kind of insight, as well as to understand to what extent the neural net recapitulates known properties of the mammalian visual pathway.  We have carried out this analysis, and the results are reported in Section \ref{findings}.  We show that our findings are quite consistent with the data analytic results on image patches in natural images obtained in \cite{klein}.  In addition, we are able to study the learning process in one example, and also to study a very deep pre-trained neural network, with interesting results which clarify the roles played by the different layers in the network.

Having performed these experiments, we became interested in the question of how to apply the knowledge obtained from our study to deep learning more generally.  In particular, we asked how one might generalize the convolutional neural net (CNN)  construction to other data sets, so as to obtain methods for constructing efficient nets that are well adapted to other large classes of  data sets, or individual data sets. We found that the key idea from the image CNN construction is the fact that the set of features (pixels) is endowed with a geometry, which can be encoded in a metric, coming from the grid in which the pixels are usually arranged.  However, in most data sets, one has one or more natural notions of distance between features, and generalizations based on such metrics appeared to be a potentially very powerful source of methods for constructing neural nets with restrictions on the connections based on such a metric. The idea of studying geometric properties of features has been foreseen by M. Robinson in \cite{robinson} under the heading of {\em topological signal processing}.  The second goal for us in this paper, then, is to introduce a mathematical formalism for constructing neural network structures from metric and graph based information on the feature space of a data set.
We also find that this formalism simplifies and makes precise the specification of neural networks even while using standard methods.     In Section \ref{dataanalytic} we  evaluate the improvements possible from  the very simplest application of this idea.  The nature of the improvements come in two directions.  The first is in speeding up the learning process.  The training of neural nets can be quite a time consuming process, and it is clearly desirable to lower the cost (in time) of training.  We found that the methods were more effective on more complex data sets, which is encouraging.  A second kind of improvement is in the direction of {\em generalization}.  When training on image data sets, it is standard procedure to select two subsets of the data set, one the training set and the other the test set.  The network is trained on the training set, and accuracy is evaluated on the test set.  This procedure is designed to guard against overfitting, and the accuracy often achieves very impressive numbers.  However, one can consider the problem of training on one data set of images and evaluating on an entirely different data set.  For example, there are two familiar data sets consisting of images of digits, one MNIST \cite{mnist} and the other SVHN \cite{svhn}.  The first is a relatively ``clean" data set, The second is actually obtained from images of numbers for the addresses of houses.  One could attempt to train on MNIST and evaluate accuracy  not on a different subset of MNIST, but rather SVHN.  Surprisingly, the results of this process yield abysmal results, with an accuracy very close to that achieved by random selection of classifications.  We demonstrate that by the use of the methods we have discussed one can improve the accuracy significantly, although still not to an acceptable level.  It suggests that further application of the methods could give us much improved generalization.  

We identify three separate scenarios giving rise to geometric information about the feature space.  The first is where by its very construction, a set of features is equipped with a geometric structure.  Typical examples of this situation are images or time series, where, for example, the pixels (features of images) are designed with a rectangular geometry in mind.  The second is where a geometry is obtained from   studies such as that performed in \cite{klein}.  Finally, there is a situation where one is given a more or less general data matrix with numerical entries, and imposes a metric  on it via standard choices of metric such as Euclidean, Hamming, etc.  Once this has been done, it is important to be able to compress this geometric information into a smaller representation, something which can be achieved by the Mapper construction \cite{mapper}.  

We believe that the study of the geometry of the feature space attached to various kinds of data sets will be a very powerful tool that can inform the construction  and improve the performance of neural networks.  Additionally, because we have incorporated geometric methods in the constructions, we also believe that our formalism opens the door to more sophisticated, detailed, and nuanced mathematical analysis of neural networks.  

\section{Neural Nets} \label{neuralnets}
This section will introduce feed-forward neural nets as well as the special case of convolutional neural nets (CNN's).  
\begin{Definition} By a {\em feed-forward system} of depth $r$ we will mean a directed acyclic graph $\Gamma$ with vertex set $V(\Gamma)$ with the following properties.
\begin{enumerate}
\item{$V(\Gamma)$ is decomposed as a disjoint union 
$$  V(\Gamma) = V_0(\Gamma) \sqcup V_1(\Gamma) \sqcup \cdots \sqcup V_r(\Gamma)
$$ }
\item{If $v \in V_i(\Gamma)$, then every edge of the form $(v,w) $ of $\Gamma$ has $w \in V_{i+1}(\Gamma)$.  }
\item{The nodes in $V_0(\Gamma)$ (respectively $V_r(\Gamma )$) are called {\em initial nodes }(respectively {\em terminal nodes}).  }
\item{We assume that for every non-initial node $w \in V_i(\Gamma)$, there is at least one $v \in V_{i-1}(\Gamma)$ so that $(v,w) $ is an edge in $\Gamma$.}
\item{ For each vertex $v$ of $\Gamma$, we denote by $\Gamma(v)$ (respectively $\Gamma ^{-1}(v)$) the set of all vertices $w$ of $\Gamma$ so that $(v,w)$ (respectively $(w,v)$)  is an edge of $\Gamma$.  }
\end{enumerate} 
The sets  $V_i(\Gamma )$ are referred to as the {\em layers} of the feed-forward system.  We say that a layer $V_i(\Gamma)$  is {\em locally finite} if the sets $\Gamma ^{-1}(v)$ are finite for all $v \in V_i(\Gamma )$.  By a {\em sub-feed-forward system} of a feed-forward system $\Gamma$ of depth $r$, we mean a directed subgraph $\Gamma _0 \subseteq \Gamma$ so that the graph $\Gamma _0$ and the families of vertices $V_0(\Gamma ) \cap \Gamma_0, \ldots , V_r(\Gamma ) \cap \Gamma _0$ themselves form a feed-forward system.   In particular, it must be the case that for each $v \in \Gamma _0$, the set $\Gamma ^{-1}(v) \cap \Gamma _0$ must be non-empty.  

\end{Definition} 
\begin{Remark} {\em Note that we do not assume that $\Gamma$ is finite. It is sometimes useful to use infinite feed forward systems  as  idealized constructions with useful finite systems  contained in it.  }
\end{Remark} 
\begin{Remark}{\em We have described only the simplest kinds of structures used in neural nets.  There are many others, which can also be described using the methodology we are introducing, but we leave them to future work.  }
\end{Remark} 
It is also useful to have a slightly different point of view on feed-forward systems.  Recall that a correspondence from a set $X$ to a set $Y$ is a subset ${\cal C} \subseteq X \times Y$.  It is clear that one can compose correspondences, and for any correspondence ${\cal C}:X \rightarrow Y$ we will write ${\cal C}(x) = \{y \in Y|(x,y) \in {\cal C}\}$ and ${\cal C}^{-1}(y) = \{ x  \in X |(x,y) \in {\cal C}\}$. We also say that a correspondence ${\cal C}: X \rightarrow Y$ is {\em surjective} if ${\cal C}^{-1}(y) \neq \emptyset$ for all $y \in Y$.   These notions are  familiar, but we give some particular  examples that will be  relevant for the construction of convolutional neural networks.  

\begin{Example} \label{complete}{\em Given any two sets $X$ and $Y$, we have the {\em complete correspondence} ${\cal C}^c(X,Y) : X \rightarrow Y$, defined b y ${\cal C}^c(X,Y) = X \times Y$.  }
\end{Example} 
\begin{Example}{\em Given any map of sets $f:X \rightarrow Y$, we have the {\em functional correspondence} ${\cal C}_f: X \rightarrow Y$ attached to $f$, defined to consist of the points in the graph of $f$, defined to be  $ \{ (x,f(x))|x \in X \} $.    }
\end{Example} 
\begin{Example}\label{product}{\em  Let ${\cal C}:X\rightarrow U$ and ${\cal D}: Y \rightarrow W$, we define the {\em product correspondence} 
$$ {\cal C} \times {\cal D} : X \times Y \rightarrow U \times W
$$
by the requirement that $((x,y)(u,w)) \in {\cal C} \times \cal {D}$ if and only if $(x,u) \in {\cal C}$ and $(y,w) \in {\cal D}$. }
\end{Example} 
\begin{Example}\label{metriccorr} {\em Let $X$ be a metric space, with distance function $d$.  Suppose further that we are given a non-negative threshold $r$.  Then we define ${\cal C}_d(r): X \rightarrow X$, the {\em metric correspondence with threshold $r$} from $X$ to itself, by ${\cal C}_d(r)(x) = \{ x^{\prime} | d(x,x^{\prime})  \leq r \}$. It will occasionally be useful to permit the definition of metric spaces to include the possibility of infinite values.  The three axioms of metric spaces extend in a natural way to this generality.   }
\end{Example} 
\begin{Example}\label{graphcorr} {\em Let $\Gamma$ be graph, with vertex set $V = V(\Gamma )$.  Then the {\em graphical correspondence}  ${\cal C}_{\Gamma} : V \rightarrow V$ is defined by $(v,v^{\pr}) \in {\cal C}_{\Gamma}$ if and only if $(v,v^{\pr})$ is an edge in $\Gamma$.   }
\end{Example} 

We now give the definition of a  kind of object that is completely equivalent to a feedforward system.

\begin{Definition} \label{generator} Let ${\cal I}_r$ denote the totally ordered set $\{ 0, 1, \ldots , r \}$ regarded as a category.   By a {\em generator} for an $r$-layer feed-forward system, we will mean a functor $F$ from the  category ${\cal I}_r$ to the category $\underline{Cor}$ of finite sets and correspondences.  The associated feed-forward system has as its vertex set $\coprod F(i)$, and where there is a connection from  $v \in F(i)$ to $w \in F(j)$ if and only if (1) $j = i+1$ and (2) $(v,w) \in F(i \rightarrow i+1)$.   \end{Definition}

Feed-forward systems are used to describe and specify  certain computations.  The nodes are considered variables, so will be assigned numerical values which we call $r_{v}$.    The nodes in the $0$-th or initial layer are regarded as input variables, so they are in one to one correspondence with variables that are attached to a data set. 

\begin{Definition}  By an {\em activator}, we will mean a triple  $(\mu, S,f)$, where $\mu$ is a commutative semigroup structure on $\bb{R}$, $S$ is a subsemigroup of the multiplicative semigroup of  $\bb{R}$, and $f:\bb{R} \rightarrow \bb{R}$  is a function, which we call the {\em cutoff function}.  Given a feedforward structure $\Gamma$, an {\em activation system} for $\Gamma$ is a choice of an activator $(\mu _{v}, S_v,f_v)$ for each non-initial vertex of $\Gamma$.  A {\em coefficient system} for a feed-forward system $\Gamma$ and activation system $(\mu _v , S_v, f_v)$ is a choice of element $\lambda_{(u,v) } \in S_{v}$ for each edge $(u,v)$ of $\Gamma$.  
\end{Definition} 
\begin{Remark} {\em  Typically  we use only a small number of distinct activators, and also assign all the nodes in a given layer the same activator.  For purposes of this paper, the only semigroup structures on $\Bbb{R}$ we use are the additive structure and the commutative operation $(x,y) \rightarrow \mbox{max}(x,y)$.  Also, for the purposes of this paper, the only choices for $S$ will be either all of $\bb{R}$ or $\{ 1 \}$, but in other contexts there might be other choices. The cutoff function may  be chosen to be the identity, but in general is a continuous function that is a continuous version of a function that is zero below a threshold and 1 above it.   The ring $\bb{R}$ can be replaced by other rings, such as the field with two elements, which can be useful in Boolean calculations.   }
\end{Remark}

We now wish to use this data to construct functions on the input data. We assume we are given a locally finite feed-forward structure $\Gamma$, equipped with an activation system ${\cal A} = (\mu _v,S_v, f_v)$ and a coefficient system $\Lambda = \{ \lambda _{(u,v)}\}$.   For each $i$, with $0 \leq i \leq r$, we set $W_i$ equal to the real vector space of functions from $V_i(\Gamma )$ to $\bb{R}$.    We now define a function $\varphi _i = \varphi_{i}(-:{\cal A},\Lambda): W_{i-1} \rightarrow W_i$, for $0 \leq i \leq r$, on a function $g:V_{i-1} \rightarrow \bb{R}$ by 
$$ \varphi_i(g)(v) = f_v(\sum _{(u,v)\in \Gamma } \lambda_{(u,v)} g(u))
$$
Note that the sum is computed using the monoid structure $\mu _v$, and is taken over all edges of $\Gamma$ with terminal vertex $v$.  This set is finite by the local finiteness hypothesis.  We have now constructed functions $\varphi _i: W_{i-1} \rightarrow W_i$ for all $ 0 \leq i \leq r$, and therefore can construct the composite 
$$ \Phi = \Phi (-;{\cal A},\Lambda) = \varphi _r \compcirc \varphi _{r-1} \compcirc \cdots \compcirc \varphi _1
$$
from $W_0$ to $W_r$, i.e. a function from the input set to the output set.  

The final requirement is the choice of a {\em loss function}.   Given a set of points 
$\frak{D} \subseteq W_0$, and a function $F:\frak{D} \rightarrow W_r$, the goal of deep learning methods is to construct a function $\varphi$ as above that best approximates the function $F$ in a sense which has yet to be defined. If the function is viewed as a continuous function to the vector space $W_r$, then the finding the best $L_2$ approximation is quite reasonable, and the $L_2$ distance from the approximating function to $F$ will be defined to be the loss function.  If, however, the output function is categorical, i.e. has a finite list of values, then it is often the case that  the possible outputs are identified with the vertices in the standard $(n-1)$-simplex 
$$\{ (x_1, \ldots, x_n)| x_i \geq 0 \mbox{ for all } i \mbox{ and } x_1 + \cdots + x_n = 1 \}
$$
 in $\bb{R}^n$, and other loss functions are more appropriate.  The output function  still takes continuous valued, and  the goal becomes to fit a continuous function to the discrete one.  One could of course do this directly, but it has been found that fitting certain transformations of the continuous function  perform better.  One very common choice is the following.  Suppose that from the construction of the neural net, it is known that the values of the neurons in the terminal layer are always positive real numbers.  Define $\sigma_n: \bb{R}_+ ^n \rightarrow \bb{R}_+^n$ by 
\begin{equation} \label{sigma} \sigma_n (x_1, \ldots , x_n ) = \frac{1}{x_1 + \cdots + x_n} (x_1, \ldots , x_n)
\end{equation}
The function $\sigma _n$ takes its values in the standard $(n-1)$  simplex. 
The {\em softmax} function is the composite $\sigma \compcirc \mbox{exp}$, where $\mbox{exp}$ denotes the function $(x_1, \ldots , x_n) \rightarrow (e^{x_1}, \ldots , e^{x_n})$ from $\bb{R}^n $ to $\bb{R}^n$.  A standard procedure for optimizing fitting a continuous function $F$  with discrete values $\{\alpha _1, \alpha _2, \ldots \alpha _n \}$  is to minimize  the $L_2$ error of the transformed function 
$$  \sigma_n \compcirc exp \compcirc \varphi 
$$
where $n$ is the number of neurons in the output layer.  
This notion of loss or  error is referred to as the softmax loss function.

Deep learning proceeds to minimize the  chosen loss function  of  the difference between $\Phi (-;{\cal A},\Lambda) $ and a given function $g$ over the possible choices of the coefficients $\lambda _{(v,w)}$ using a stochastic variant of the gradient descent method.   Note that $F$ is typically empirically observed, it is not given as a formula.  The optimization process often is time consuming, and occasionally becomes stuck in local optima.  We refer to a feed-forward system equipped with activation system as a {\em neural net}. 

\begin{Definition} Consider a locally finite feed-forward system $\Gamma$, possibly infinite, equipped with an activation system ${\cal A}$.    Let $\Gamma _0 \subseteq \Gamma$ be a sub-feed-forward system.  If ${\cal A}$ is an activation system on $\Gamma$, then it is clear that its restriction ${\cal A}|\Gamma _0$ to $\Gamma _0$ is an activation system for $\Gamma _0$ and that similarly,  a coefficient system $\Lambda $ on $\Gamma$ restricts to an coefficient system $\Lambda |\Gamma _0$ on $\Gamma _0$.  We will call the neural net $(\Gamma _0, {\cal A}|\Gamma _0) $ the {\em restriction} of the neural net $(\Gamma , {\cal A})$ to $\Gamma _0$.  
\end{Definition}

There is an additional kind of structure on a feed-forward system that is particularly useful for data sets of images, as well as other types of data. 

\begin{Definition} \label{defcon} By a {\em convolutional structure } on a layer $V_i(\Gamma )$ in a feed-forward system $\Gamma$  we mean a pair $(\simeq, \psi )$, where $\simeq $ is an equivalence relation on the set of  vertices  of $V_i(\Gamma) $, and where $\psi$ is  an assignment of a bijection 
$$\psi _{(v,w)}: \Gamma ^{-1}(v) \rightarrow \Gamma ^{-1}(w)
$$
for any pair $(v,w)$ in $\simeq$,
 satisfying the requirement that  $\psi _{(v,w)} = \psi _{(w,v)} ^{-1}$ and $\psi_{(w,v)} = \psi _{(w,u)} \compcirc \psi _{(u,v)}$ when defined.  An activation system for $\Gamma$ is said to be {\em adapted} to the convolutional structure on a layer $V_i(\Gamma)$ if whenever $v \simeq w$, it is the case that $(\mu_v, S_v, f_v) = (\mu _w, S_w, f_w)$.
 A  coefficient system  $\{ \lambda _{(v,w)}\}$ for the neural net $(\Gamma, {\cal A})$ is {\em adapted} to a convolutional structure
$\{ \simeq, \psi_{(v,w)}\}$ if it satisfies the compatibility requirement that whenever $v \simeq w$, then we have 
$$ \lambda_{(u,v)}  = \lambda_{(\psi _{(v,w)}(u),w)}
$$
for all $u \in \Gamma ^{-1}(v)$. 
\end{Definition}

\begin{Example}\label{convdef} {\em  Suppose that a layer $V_i(\Gamma)$ and the layer $V_{i-1}(\Gamma)$ are acted on by a group $G$, and suppose further that for any $v \in V_{i-1}(\Gamma)$ and $w \in V_i (\Gamma )$, 
$(v,w)$ is an edge in $\Gamma$ if and only if $(gv,gw)$ is an edge for all $g \in G$.        Suppose further that the actions on both $V_{i-1}(\Gamma)$ and $V_i(\Gamma )$ are free, so that the only element  of $G$  that fixes a node is the identity element. We define an equivalence relation $\simeq$ on  $V_i(\Gamma)$ by declaring that $ v \simeq w$ if and only if there is an element $g \in G$ so that $gv = w$. Because of the freeness of the action, $v$ and $w$ determine $g$ uniquely.  We define the bijection $\psi_{(v,w)} : \Gamma ^{-1}(v) \rightarrow \Gamma ^{-1}(w)$ to be multiplication by $g$.  Because the group preserves the directed graph structure in $\Gamma$, $g$ does carry 
$\Gamma ^{-1}(v)$ to $\Gamma ^{-1}(w)$.  The application of this idea to data sets of images uses the group $\bb{Z}^2$, whose points correspond to an infinite pixel grid.   We call structures defined this way {\em Cayley structures}.  }
\end{Example}

The  description of a convolutional layer in Example \ref{convdef} is useful in many situations where the group, and therefore the feed-forward system, are infinite.  Nevertheless, it is useful to adapt the networks to finite regions in the grid, such as $N \times N$ grids within an infinite pixel grid.  This fact motivates the following definition.  

\begin{Definition} \label{restriction}
 We suppose that we have a feed-forward structure $\Gamma$, a layer $V_i(\Gamma)$ equipped with  a convolutional structure $\{ \simeq, \psi _{(v,w)} \}$, and  a sub-feed-forward structure $\Gamma _0 \subset \Gamma$. The restriction of the equivalence relation to $V_i(\Gamma _0)$ does give an equivalence relation on $V_i(\Gamma _0)$, but it does not necessarily have the property that the restriction of the bijections $
\psi _{(v,w)}$ to $\Gamma ^{-1}(v)\cap V_{i-1}(\Gamma _0)$ remains a bijection.  We will define an equivalence relation $\simeq _0$  on $V_i(\Gamma _0)$ by declaring that $v \simeq _0 w$ if and only if {\em (}a{\em )}  $v \simeq w$ as vertices in $V_i(\Gamma)$ and {\em (}b{\em ) } $\psi _{(v,w)} $ restricts to a bijection from $\Gamma ^{-1}(v) \cap V_{i-1}(\Gamma _0)$ to $\Gamma ^{-1}(w) \cap V_{i-1}(\Gamma _0)$.  This clearly produces a convolutional structure on the layer $V_i(\Gamma _0)$ in the feed-forward structure $\Gamma _0$, which we refer to as the {\em restriction } of the convolutional structure e $\{ \simeq, \psi _{(v,w)} \}$  on $V_i(\Gamma)$ to $\Gamma _0$.    
\end{Definition}

\section{Natural Images and Convolutional Neural Nets} \label{images}  Data sets of images are of a great deal of interest for many problems.  For example, the task of recognizing hand drawn digits or letters from images taken of them is a very interesting problem, and an important test case.  Neural net technology has been successfully applied to this situation, but in many ways the success is not well understood, and it is believed that it is often due to overfitting.  Our goal is to understand the operation of this methodology  better, and to use that understanding to improve performance in terms of speed, and of the ability to generalize from one data set to another.  In this section we will discuss image data sets, the feed-forward systems that have been designed specifically for them, the extent to which the neural networks act similarly to learning in humans and primates, and how such insights can be used to speed up and improve generalization from one image data set to another.  

By an {\em image}, we will mean an assignment of numbers (gray scale values) to each pixel of a pixel array, typically arranged in a square. The image can be regarded as a $P$-vector, where $P$ denotes the number of pixels in an array.  However, the grid structure of the pixels  tells us that there is additional information, namely a geometric structure on the set of coordinates in the vector.  It turns out to be useful to build neural nets with a specific structure, reflecting this fact.  For simplicity of discussion, it turns out to be useful to build infinite but locally finite models first, and then realize the actual computations on a finite subobject of these infinite constructions, by restricting the support of the activation systems we consider in the optimization.   We will be specifying our neural networks by generators.  First, we let $\bb{Z}$ denote the integers.  By $\bb{Z}^2 = \bb{Z} \times \bb{Z}$ we will mean the metric space whose elements consist of ordered pairs of integers, and where the distance function is the restriction of the $L^{\infty}$ distance on   $\bb{R}^2$.  We of course have the metric correspondences from $\bb{Z}^2$ to itself.  We will define another family of correspondences called {\em pooling correspondences}. For any pair of integers $m \leq n$, let $[m,n]$ denote the intersection of the interval $[m,n]$ in the real line with the integers.  Let $N$ denote a positive integer, and define a correspondence $\pi (m,n,N)$ to be $\alpha ^{-1}$ where $\alpha: \bbz \rightarrow \bbz$ is defined by  $\alpha (x) = 
[Nx +m, Nx + n]$.  We have two parameters that are of interest for these correspondences, the {\em stride}, which is the integer $N$, and the {\em width}, which is the integer $n-m+1$.  To give a sense of the nature of these correspondences, consider the situation with stride and width both equal to 2, and with $m = 0$. In this case, it is easy to check that the correspondence $\pi(0,1,2)$ is given by $x \rightarrow \lfloor \frac{x}{2} \rfloor$.  In general,  if the stride is equal to the width, the correspondence $\pi (m,n,N)$ is actually functional, and the corresponding function is $N$ to $1$.  
 We'll write $\pi ^{s}(m,n,N)$ for the $s$-fold product of $\pi(m,n,N)$ as a correspondence from $\bb{Z}^s$ to itself.  

It will be useful to have a language to describe the layers in a feed-forward system in terms of the generators.  

\begin{Definition}  Let $\Gamma $ denote a feed-forward system, with generator $F:{\cal I}_r \rightarrow \underline{Cor}$.  For any $i \in {\cal I}_r$, we consider the $i$-th layer $F(i)$ as well as the correspondence $\theta _i = F(i-1 < i): F(i-1) \rightarrow F(i)$. 

\begin{enumerate} 
\item{We say the layer $F(i)$ is {\em fully connected} if $\theta _i$ is the complete correspondence ${\cal C}^c(F(i-1),F(i))$, as defined in Example \ref{complete}.    }
\item{We say $F(i)$ is {\em grid convolutional} if there are sets $X$ and $Y$, so that $\theta _i$ is of the form 
$$  {\cal C}^c(X,Y) \times {\cal C}_d(r): X \times \bb{Z}^2 \rightarrow Y \times\bb{Z}^2
$$
where ${\cal C}_d(r)$ is  a metric correspondence as defined in Example \ref{metriccorr}. }
\item{We say $F(i)$ is {\em pooling} if $\theta _i$ is of the form  
$$  {\cal C}^c(X,Y) \times \pi ^2(m,n,N): X \times \bb{Z}^2 \rightarrow Y \times\bb{Z}^2
$$}
\end{enumerate}  
\end{Definition} 

\begin{Remark}{\em The reason for taking the product of convolutional or pooling correspondences with complete correspondences is in order to accommodate the idea of including numerous copies of a grid within a layer, but with the understanding that the graph connections between any copy of a grid  in $F(i-1)$ and any copy in $F(i)$ are identical.  This is exactly what the product correspondence achieves.  }
\end{Remark}


We are now in a position to build some convolutional neural networks. We will do so by constructing a generator.  The generator is a functor that can be specified by a diagram like the following, where writing $X(n)$ denotes a set of cardinality $n$.  

\begin{equation} \label{elementary} X(1) \times \bb{Z}^2  \xrightarrow{\footnotesize\makebox[1.8cm]{${\cal C}^c \times {\cal C}_d(1)$}} X(64) \times \bb{Z}^2
 \xrightarrow{\footnotesize\makebox[2.3cm]{${\cal C}^c \times \pi^2(0,1,2)$}} X(64) \times \bb{Z}^2 
  \xrightarrow{\footnotesize\makebox[.5cm]{${\cal C}^c $}} X(10)
  \end{equation}
  
  To further simplify the description, we note that there is product decomposition of the functor $F$.  For an two  functors $F,G : \underline{C} \rightarrow \underline{Cor}$, we can form  the product functor $F \times G$, which is defined to be the point wise product on object, and which also forms the product correspondences.  It is clear from the description above that the functor we have described decomposes as the functor $F_0 \times F_1$, where $F_0$ is given by 

$$  X(1) \stackrel{{\cal C}^c}{\longrightarrow} X(64) \stackrel{{\cal C}^c}{\longrightarrow}  X(64) \stackrel{{\cal C}^c} {\longrightarrow }X(1)
$$

and  $F_1$  by 
$$ \bb{Z}^2  \xrightarrow{\footnotesize\makebox[1.0cm]{${\cal C}_d(1) $}} \bb{Z}^2 
 \xrightarrow{\footnotesize\makebox[1.8cm]{$\pi^2(0,1,2) $}} \bb{Z}^2 
  \xrightarrow{\footnotesize\makebox[.5cm]{${\cal C}^c $}} X(10)
$$ 

This kind of decomposition is ubiquitous for neural networks, where there is one functor $F$ consisting entirely of complete correspondences.  We will say a generator $F$  is {\em complete} if each of the correspondences $F(i < i+1)$ is a complete correspondence, and describe generators $F$ as $F = F^{c} \times F^s$, where $F^c$ is a complete correspondence, and $F^s$ will be referred to as the {\em structural generator}.  We note that a complete correspondence $F$ is completely determined by the cardinalities of the sets $F(i)$, and so we specify $F$ by its list of cardinalities.  We say that the {\em type} of a complete generator  $F:{\cal I}_r \rightarrow \underline{Cor}$ is the list of integers 
$$ [\#F(0), \#F(1), \ldots , \#F(r) ]
$$
and note that the type determines the structure of $F$.


\section{Findings}\label{findings}  Because of the stochastic nature of the optimization algorithms used in convolutional neural nets, the problem of understanding how they function is a problem in data analysis.  What we mean by this is that it is a computational situation where there are outliers which are not meaningful, and a useful analysis must proceed by understanding what the most common (or dense) phenomena are, in a way that permits one to ignore the outliers, which will be sparse. Before diving into the methodology and results of our study, we will talk about  earlier work \cite{klein} on the statistics of natural images which is quite relevant to our results on convolutional neural nets. 

The work in \cite{klein} was a study of a data set constructed by Mumford et al in \cite{mumford} based on a database of images collected by van Hateren and van der Schaaf \cite{hateren}.  The images were taken in Groningen  in the Netherlands, and Mumford et al collected a data set consisting of $3 \times 3$ patches, thresholded from below by variance of the patch.  Each patch consists of nine gray scale values, one for each pixel.  The data was then mean centered, and the contrast (a weighted version of variance) was normalized to have value 1.  This means that the data can be viewed as residing on the sphere $S^7$, a subspace of $\Bbb{R}^8$.  Finally, the data was filtered by {\em codensity}, a function on the data set defined at a point $x$ to be the distance from $x$ to its $k$-th nearest neighborhood.  The integer $k$ is a parameter, much as variance is a parameter in kernel density estimators, and the codensity varies inversely with density.

What was done in \cite{klein} as to select a threshold value $\rho$ (a percentage) for the codensity computed for a value $k$, and consider only points whose codensity was less than $\rho$.  For example,  one might study the set of data points which are among the lowest $25 \%$ in codensity, computed for the parameter value $k = 300$.  This was carried out in \cite{klein} for a $30 \%$ threshold value, and for the parameter values $k = 300$ and $k = 15$. 

{\begin{figure}[!htp]
\centering
\begin{minipage}[t]{0.4\textwidth} 
\includegraphics[width=\textwidth]{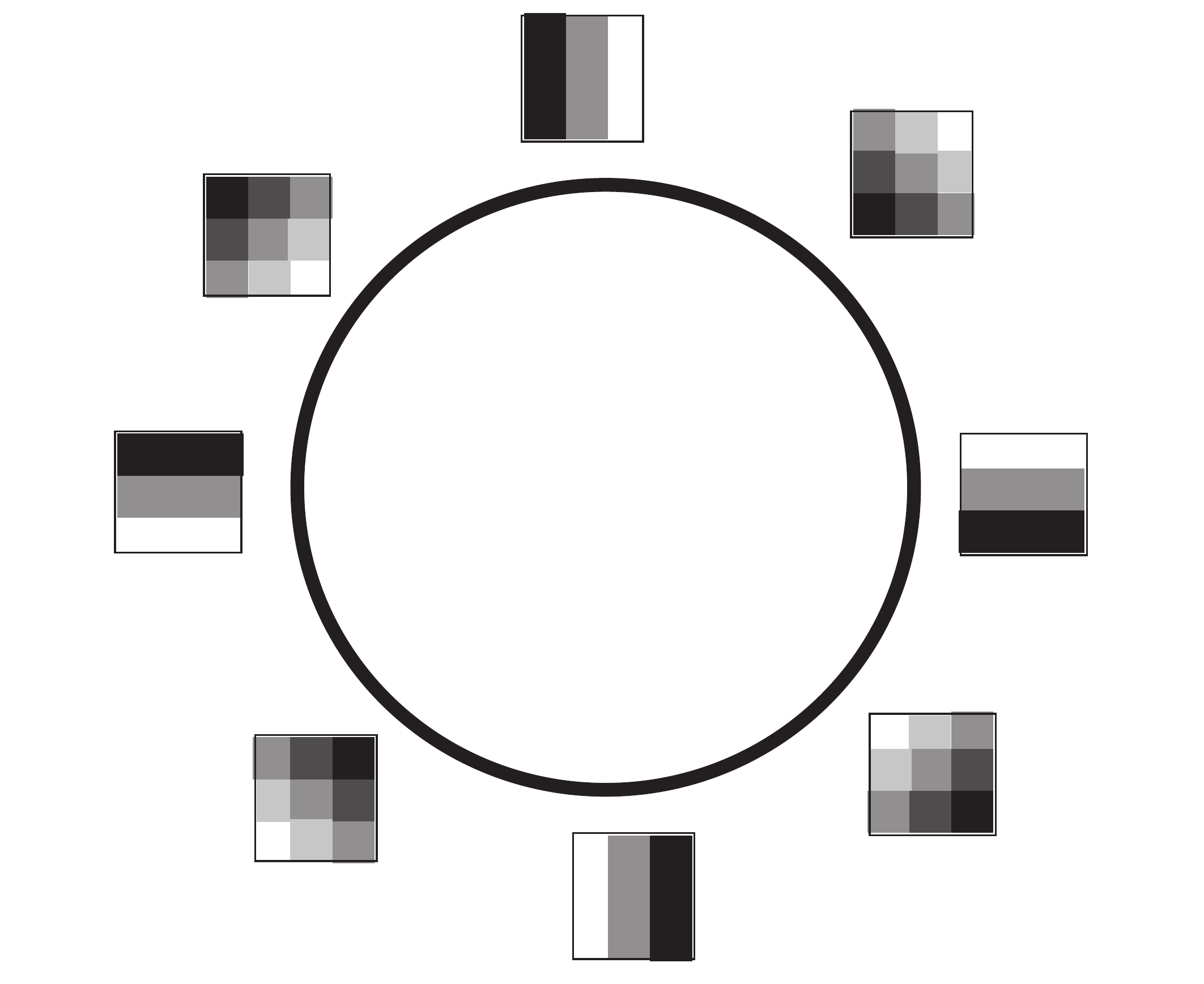}
\caption{$k=300, \rho = 30 \%$}\label{here}
\end{minipage} 
\begin{minipage}[t]{0.4\textwidth}
\includegraphics[width=\textwidth]{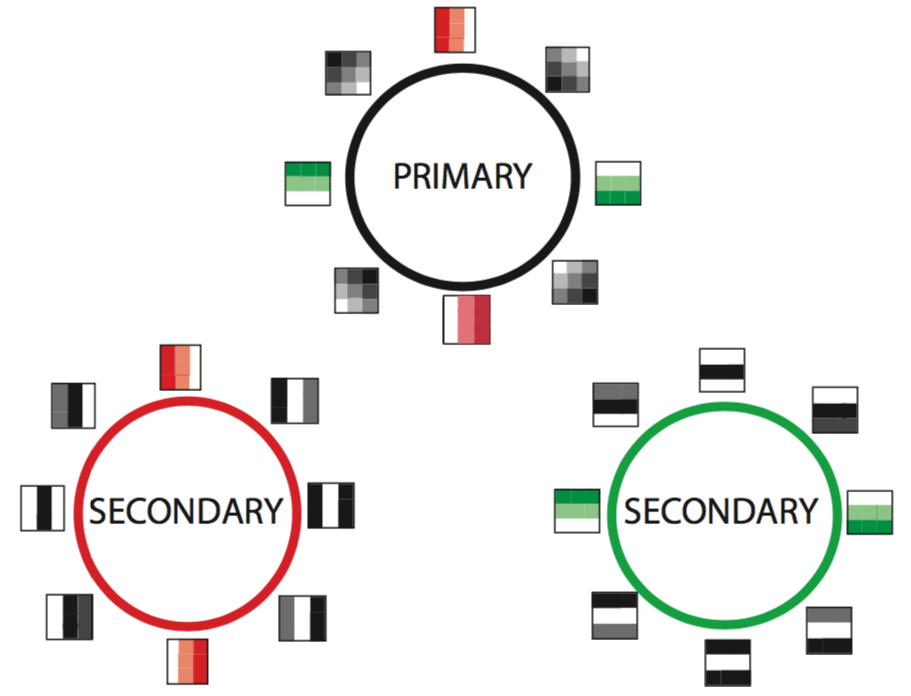}
\caption{$k=15,  \rho = 30 \%$} 
\label{three}
\end{minipage}
\end{figure}}
These diagrams were obtained by examining the data following persistent homology computations which showed $\beta _1 = 1$ in the case of Figure \ref{here} and $\beta _1 = 5$ in the case of Figure \ref{three} (note that in the case of Figure \ref{three} the model is not actually three disjoint circles, instead each of the secondary circles intersects the primary circle in two data points.  The work in \cite{klein} went further and found more relaxed thresholds that yielded a Klein bottle instead of just a one skeleton, indicating that more is going on. It  meant that the data set actually included arbitrary rotations of the two secondary circles in Figure \ref{three}.  The original motivation for the work in \cite{hateren} and \cite{mumford} was to understand if analysis of the spaces of patches in natural images is reflected in the ``tuning" of individual neurons in the primary visual cortex.  
We set out to determine if the statistical analysis of \cite{klein} has a counterpart in  convolutional neural networks  for the study of images.  The following are  insights  we have obtained.
\begin{itemize}
\item{The role of thresholding by density or proxies for density is crucial in any kind analysis of this kind.  Without that a very small number of outliers can drastically alter the topological model from something giving insight to something essentially useless.  }
\item{The development of neural networks was based on the the idea that neural networks are  analogous to networks of neurons  in the brains of mammals.   There is an understanding \cite{hubelwiesel} that the primary visual cortex acts as a detector for edges and lines, and also that higher level components of the visual pathway detect more complex shapes.  We perform an analysis analogous to the one in \cite{klein}, and show that it gives results consistent with the density analysis performed there. }
\item{We demonstrate that our observations can be used to improve the ability of a convolutional neural network to generalize from one data set to another.  }
\item{We demonstrate that the results can be used to speed up the learning process on a data set of images}

\end{itemize}

We next describe the way that the data analysis was performed.  We suppose that we have fixed an architecture for a convolutional neural network analysis of a data set of images, using grid layers as described in Section  \ref{images}.  We used an architecture in which the correspondences ${\cal C}_d(1)$ described the connections into a convolutional layer,  where $d$ is the $L^{\infty}$ metric on the grids.  This means that any node in a grid layer is connected to the nodes which belong to a $3 \times 3$ patch surrounding it.  The weights therefore constitute a vector in $\bb{R}^9$, which corresponds exactly to raw data used in \cite{mumford}.  The data points will be referred to as {\em weight vectors}.  In \cite{gabrielsson}, we performed analyses  on data sets constructed this way using a methodology  identical to that carried out in \cite{mumford} and \cite{klein}. The rest of this section will describe the results of this study. 

We first discuss the two data sets that we studied.  The first is MNIST \cite{mnist}, which is a data set of images of hand drawn digits.  The images are given as $28 \times 28$ gray scale images. For this data set, we used an architecture described as follows. The depth is $6$, and the generator $F$  is a product of two generators, $F^{c}$ and $F^s$. The complete factor  $F^{c}$ is of type $[1,64,64,32,32, 64,1]$, and the structural factor has the form 
\begin{equation} \label{mnistnetwork} G_{28} \xrightarrow{\footnotesize\makebox[.7cm]{${\cal C}_d(1) $}} G_{28} \xrightarrow{\footnotesize\makebox[1.3cm]{$\pi ^2(0,1,2) $}} G_{14} \xrightarrow{\footnotesize\makebox[.8cm]{${\cal C}_d(1)$}} G_{14} \xrightarrow{\footnotesize\makebox[1.3cm]{$\pi ^2(0,1,2) $}} G_7 \xrightarrow{\footnotesize\makebox[.5cm]{${\cal C}^c $}}X(1) \xrightarrow{\footnotesize\makebox[.5cm]{${\cal C}^c $}}X(10)
\end{equation}
where $G_i \subseteq \bb{Z}^2$ denotes an $i \times i$ grid,  $X(i)$ denotes a set of cardinality $i$, and the output layer $X(10)$ is identified with the ten digits $0,1,\ldots , 9$.  This feed-forward structure embeds as a sub-feed-forward structure of the structure $F^s_{\infty}$ obtained by replacing all the finite grids $G_i$ with copies of $\bb{Z}^2$, into which they embed.  Therefore, the layers $F^s(1) = G_{28}$ and $F^s(3) = G_{14}$ inherit a convolutional structure from the Cayley convolutional structure (defined in Definition \ref{convdef}) on $F^s_{\infty}$, which is the convolutional structure we use.  The activation systems are defined using two different activation functions $f$.  The first is the {\em rectifier},  which denotes the function $f(x) = \mbox{max}(0,x)$, and which is often also denoted by {\em ReLU}.  The second is the identity function and the third is the exponential function $\mbox{exp}(x) = e^x$. The activation system is given on the layers $F(1)$ and $F(3)$ by $(+, \bb{R}, ReLU)$, on the layers $F(2)$ and $F(4)$ by $(\mbox{max}, \{ 1 \}, id)$, on the layer $F(5)$  by $(+,\bb{R}, ReLU)$, and on the layer $F(6)$ by $(+, \bb{R}, \mbox{exp})$. The loss function (defined on the layer $F(6)$) is the function  $\sigma _n$ defined in (\ref{sigma}) above.

We now look at results for the neural net trained on MNIST.  Figure \ref{mnist1} shows a Mapper analysis of the data set of weight vectors in the first convolutional layer in the neural net described above.  The neural net was trained 100 separate times, and each training consisted of 40,000 iterations of the gradient descent procedure.  In each node, one can form the average (in $\bb{R}^9$) of the vectors in that node.  The patches surrounding the Mapper model are such averages taken in representative nodes of the model near the given position.  We see that the Mapper model is in rough agreement with the circular model in Figure
\ref{here} above.

{\begin{figure}[!hbp]
\centering
\begin{minipage}[b]{0.40\textwidth} 
\includegraphics[width=.77\textwidth]{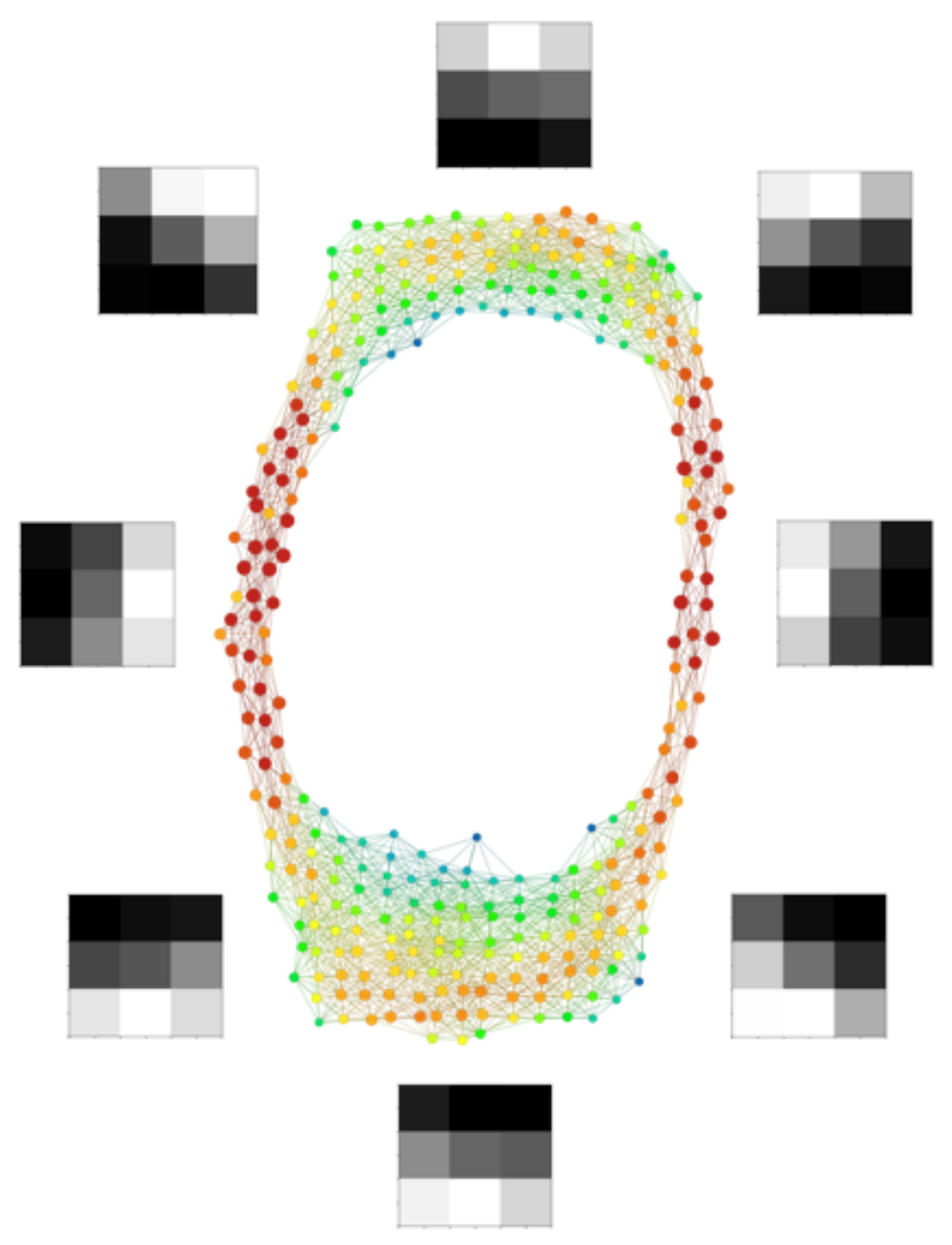}
\caption{MNIST layer 1}\label{mnist1}
\end{minipage} 
\begin{minipage}[b]{0.40\textwidth}

\center
\begin{tikzpicture}[scale = 0.60]

\draw[->] (0,0) -- (6,0) node[anchor=north] {};

\draw	(3,4) node{{\scriptsize Dimension 0}};

\draw	(0,0) node[anchor=north] {\scriptsize 0}
		(2,0) node[anchor=north] {\scriptsize 0.2}
		(4,0) node[anchor=north] {\scriptsize 0.4}
		(6,0) node[anchor=north] {\scriptsize 0.6};
\draw	(1,3.5) node{{}}
		(4,3.5) node{{}};

\draw[->] (0,0) -- (0,4) node[anchor=east] {};

\draw[->, thick, blue] (0,3.5) -- (6,3.5);
\draw[thick, blue] (0,3.1) -- (0.15,3.1);
\draw[thick, blue] (0,2.7) -- (0.10,2.7);
\draw[thick, blue] (0,2.3) -- (0.09,2.3);

\end{tikzpicture}

\begin{tikzpicture}[scale = 0.60]

\draw[->] (0,0) -- (6,0) node[anchor=north] {};

\draw	(3,4) node{{\scriptsize Dimension 1}};

\draw	(0,0) node[anchor=north] {\scriptsize 0}
		(2,0) node[anchor=north] {\scriptsize 0.2}
		(4,0) node[anchor=north] {\scriptsize 0.4}
		(6,0) node[anchor=north] {\scriptsize 0.6};
\draw	(1,3.5) node{{}}
		(4,3.5) node{{}};

\draw[->] (0,0) -- (0,4) node[anchor=east] {};

\draw[thick, blue] (0.05,3.5) -- (5.7,3.5);
\draw[thick, blue] (0.02,3.1) -- (0.38,3.1);
\draw[thick, blue] (0.32,2.7) -- (0.36,2.7);
\draw[thick, blue] (0.15,2.3) -- (0.34,2.3);
\draw[thick, blue] (0.16,1.9) -- (0.31,1.9);

\end{tikzpicture}

\hspace{1cm}
\caption{ Barcode for layer 1 } 
\label{mnist2}
\end{minipage}
\end{figure}}

In Figure \ref{mnist2}, we see persistence barcodes computed for for the data set.  The computation confirms the presence of connectedness of the data set as well as the presence of a significant loop, which is a strong indication that the Mapper model is accurately reflecting the structure of the data set.  
Figure \ref{mnist3} shows a Mapper model of the second convolutional layer.  One observes that there appear to be patches which are roughly like those in the primary circle, but the structure is generally more diffuse that what appeared in the first layer.  Persistence barcodes did not confirm a loop in this case.

{\begin{figure}[!htp]
\centering
\begin{minipage}[t]{.7\textwidth} 
\begin{center}
\includegraphics[width=.4\textwidth]{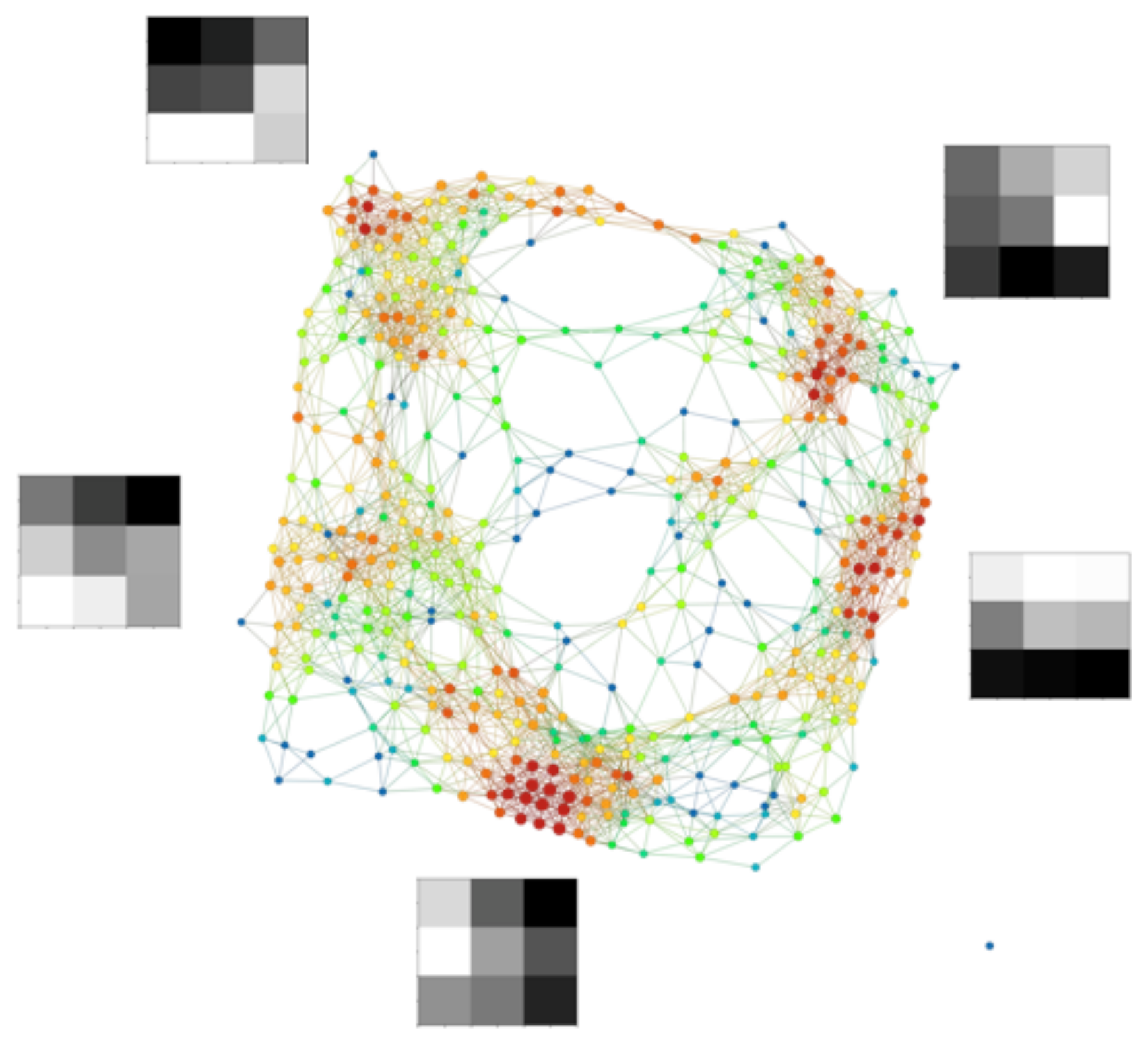}
\caption{MNIST layer 2}\label{mnist3}
\end{center}
\end{minipage} 
\end{figure}}
The second data set is CIFAR-10 \cite{CIFAR}, which is a data set of  $32 \times 32$  color images objects divided into 10 classes,  namely airplane, automobile, bird, cat, deer, dog, frog, horse, ship, and truck.  The color is encoded using the RGB system, so that each pixel is actually equipped with three coordinates, one for each of the three colors red, green, and blue.    There are different options about how to analyze color image data, and we examined three of them. 

\begin{enumerate}
\item{\label{reduce}Reduce the colors to a single gray scale value by taking a linear combination of the three color values, and then analyze the data set as a collection of gray scale images.  We used   the combination $.2989 \cdot R+ .5870\cdot G + .1140 \cdot B$. This choice is one standard choice made for this kind of problem.  See https://en.wikipedia.org/wiki/{Luma\_{\%}}28video{\%}29 for a discussion.}
\item{\label{separate} Study the individual color channels separately, producing three separate gray scale data sets, one each for red, green and blue.}
\item{\label{merge} Consider all three color channels together, and build a neural network to accommodate that.  This means in particular that the input layer will need to include three copies of the $32  \times 32$ grid.  }
\end{enumerate}

  For options (\ref{reduce}) and (\ref{separate}),  we constructed a neural net very similar to the one used for MNIST.   Its  complete factor $F^c$  is of type $[1,64,64,32,32,64,1]$, identical to the one used for MNIST.  The structural factor $F^s$ has the form 
\begin{equation} \label{cifarnetwork} G_{32} \xrightarrow{\footnotesize\makebox[.7cm]{${\cal C}_d(1) $}} G_{32} \xrightarrow{\footnotesize\makebox[1.3cm]{$\pi ^2(0,2,2) $}} G_{16} \xrightarrow{\footnotesize\makebox[.8cm]{${\cal C}_d(1) $}} G_{16} \xrightarrow{\footnotesize\makebox[1.3cm]{$\pi ^2(0,1,2) $}} G_8 \xrightarrow{\footnotesize\makebox[.5cm]{${\cal C}^c $}}X(1) \xrightarrow{\footnotesize\makebox[.5cm]{${\cal C}^c $}}X(10)
\end{equation}
The  generator is identical to the one for MNIST  except for the substitution of $G_{32},G_{16}$, and $G_8$ for $G_{28},G_{14}$, and $G_7$, respectively, and for the substitution of a pooling layer of width $3$ as the correspondence between $F^s(1) $
and $F^s(2)$.  The activation systems are identical to those in the MNIST case, as is the loss function.  
For option (\ref{merge}),  it is necessary to form an additional complete factor $G$  of type $[3,1,1,1,1,1,1]$, and form the product $F^c \times F^s \times G$ as the generator.    Of course, the $3$'s correspond to the set $\{ R,G,B \}$.  The activation systems and loss functions are identical in all three cases. 

We  first  performed an analysis in the case of option  (\ref{reduce}). The results were not as clear as in the MNIST analysis, but did give some indications of interesting structure.  In particular, the second layer had the  Mapper model shown in Figure \ref{bullseye} below.  Notice that the primary circle is included, together with a kind of ``bullseye" patch which does not appear even in the Klein bottle model given in \cite{klein}.  We also analyzed option (\ref{merge}) above.  In this case, the result was quite striking. A Mapper model of the first layer appears  in Figure \ref{threecircle}, which  we see recovers the three circle model of \cite{klein}, and where a persistence barcode for this space appears in Figure \ref{3circlebar}.  We also analyzed option \ref{separate} above, and found strong primary circles in that case.  The findings confirm that generally, the convolutional neural network well reflects the density analysis in \cite{klein}, as well as the results on the primary visual cortex given in \cite{hubelwiesel}. Moreover, the detection of the bullseye shown in Figure \ref{bullseye} demonstrates that the higher levels of the neural network find more complex patches, not accounted for by the low level analysis encoded in the Klein bottle of \cite{klein}. This is also consistent with our understanding of the visual pathway, in which there are higher level units above the primary visual cortex that capture more ``abstract" shapes. 

{\begin{figure}[!hbp]
\centering
\begin{minipage}[t]{.7\textwidth} 
\begin{center}
\includegraphics[width=.5\textwidth]{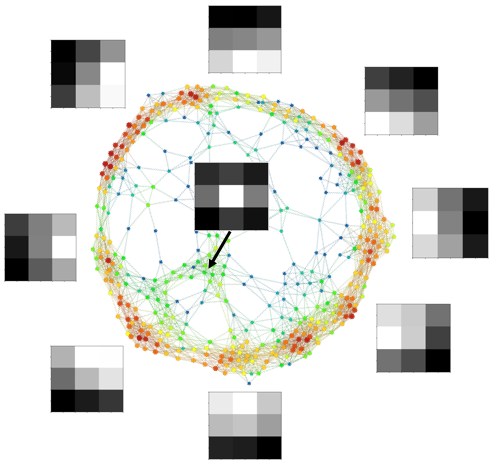}
\caption{CIFAR-10 layer 2, gray scale }\label{bullseye}
\end{center}
\end{minipage} 
\end{figure}}

{\begin{figure}[!hbp]
\centering
\begin{minipage}[b]{0.50\textwidth} 
\begin{center}
\includegraphics[width=.77\textwidth]{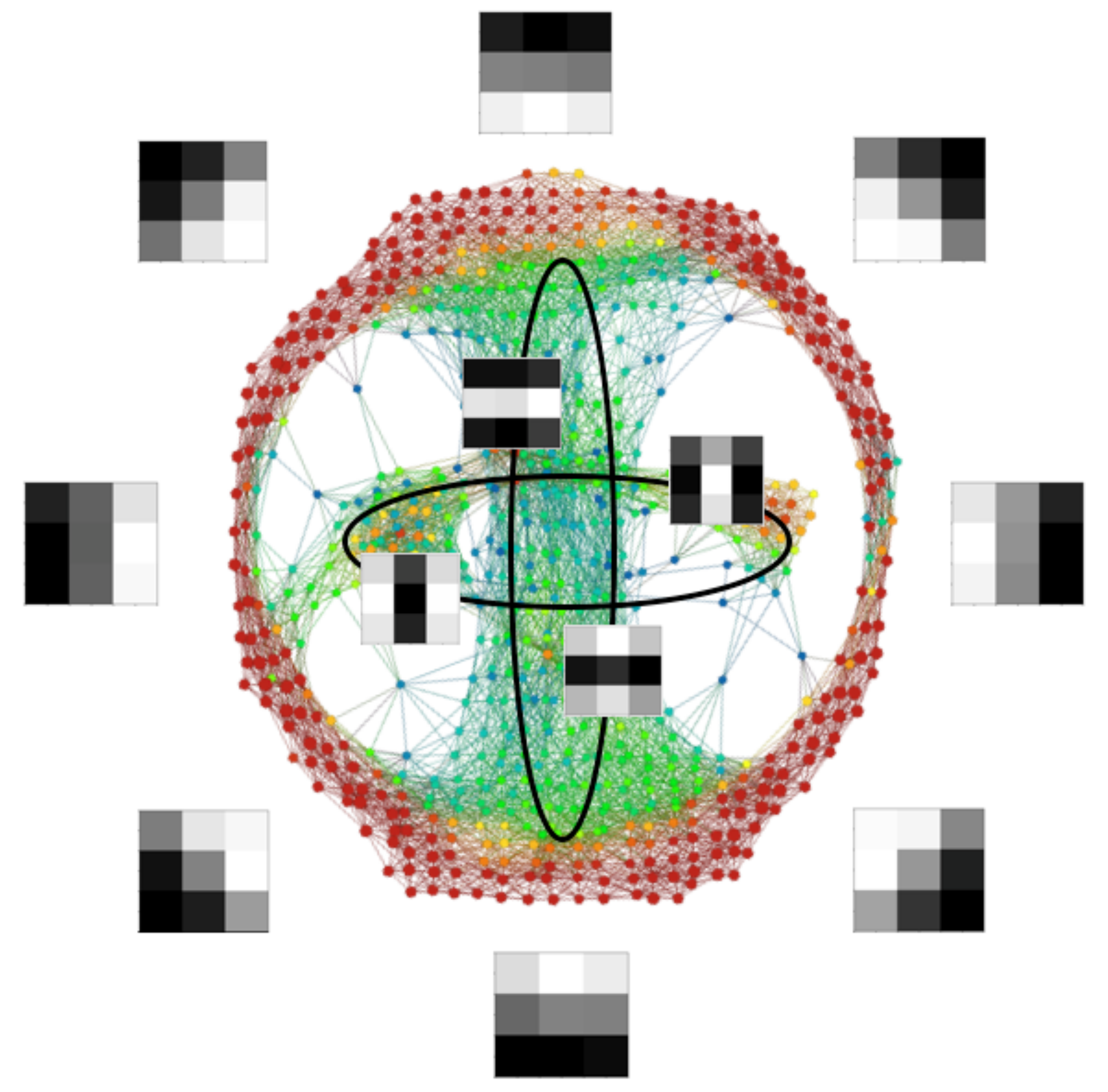}
\end{center}
\caption{First layer, CIFAR-10, separate colors }\label{threecircle}
\end{minipage} 
\begin{minipage}[b]{0.40\textwidth}
\begin{center}

\begin{tikzpicture}[scale = 0.55]

\draw[->] (0,0) -- (10,0) node[anchor=north] {};

\draw	(5,4) node{{\scriptsize Dimension 0}};

\draw	(0,0) node[anchor=north] {\scriptsize 0}
		(2,0) node[anchor=north] {\scriptsize 0.06}
		(4,0) node[anchor=north] {\scriptsize 0.12}
		(6,0) node[anchor=north] {\scriptsize 0.18}
		(8,0) node[anchor=north] {\scriptsize 0.24}
		(10,0) node[anchor=north] {\scriptsize 0.30};
		
\draw	(1,3.5) node{{}}
		(4,3.5) node{{}};

\draw[->] (0,0) -- (0,4) node[anchor=east] {};

\draw[->, thick, blue] (0,3.5) -- (10,3.5);
\draw[thick, blue] (0,3.1) -- (3,3.1);
\draw[thick, blue] (0,2.7) -- (0.7,2.7);
\draw[thick, blue] (0,2.3) -- (0.4,2.3);

\end{tikzpicture}

\begin{tikzpicture}[scale = 0.55]

\draw[->] (0,0) -- (10,0) node[anchor=north] {};

\draw	(5,4) node{{\scriptsize Dimension 1}};

\draw	(0,0) node[anchor=north] {\scriptsize 0}
		(2,0) node[anchor=north] {\scriptsize 0.06}
		(4,0) node[anchor=north] {\scriptsize 0.12}
		(6,0) node[anchor=north] {\scriptsize 0.18}
		(8,0) node[anchor=north] {\scriptsize 0.24}
		(10,0) node[anchor=north] {\scriptsize 0.30};
		
\draw	(1,3.5) node{{}}
		(4,3.5) node{{}};

\draw[->] (0,0) -- (0,4) node[anchor=east] {};

\draw[thick, blue] (0.5, 3.5) -- (8.6, 3.5);
\draw[thick, blue] (0.4, 3.1) -- (8.4, 3.1);
\draw[thick, blue] (0.1, 2.7) -- (8.4, 2.7);
\draw[thick, blue] (4.7, 2.3) -- (8.3, 2.3);
\draw[thick, blue] (4.6, 1.9) -- (6.7, 1.9);
\draw[thick, blue] (0.1, 1.5) -- (1.0, 1.5);
\draw[thick, blue] (0.7, 1.1) -- (0.8, 1.1);
\draw[thick, blue] (0.1, 0.7) -- (0.8, 0.7);
\draw[thick, blue] (0.2, 0.3) -- (0.7, 0.3);

\end{tikzpicture}

\end{center}
\caption{Persistence barcode, Figure \ref{threecircle}} 
\label{3circlebar}
\end{minipage}
\end{figure}}

We also examined the learning process for CIFAR-10.  We did this by performing the analysis in the case of option (\ref{reduce}) above at various stages of the optimization algorithm.  Figure \ref{learning} shows the results for both first and second layers.  The numbers below the models show the number of iterations corresponding to the models above them.  Most of the models shown are ``carpets", which simply reflects the choice of two filter functions for the model.  This means that they are not topologically interesting by themselves.  However, each node in the a Mapper model consists of a collection of data points, and the cardinality of that set becomes a function on the set of vertices of the model.  Sub- or superlevel sets of that function can then give interesting information, loosely correlated with density.  The models in Figure \ref{learning} illustrate this, particularly strongly in the first layer.  
We note that the first layer, beginning with something near random after 100 iterations, organizes itself into a recognizable primary circle after 200 iterations, remains at that structure until roughly 900 iterations, when the circle begins to ``degrade", and instead form a structure which is capturing patches more like those of the secondary circles.  The second layer, on the other hand, is not demonstrating any strong structure until it has undergone 1000 or 2000 iterations, when one begins to see the primary circle appearing.  One could interpret this as a kind of compensation for the changes occurring in the first layer.  
{\begin{figure}[!hbp]
\centering
\begin{minipage}[t]{.7\textwidth} 
\begin{center}
\includegraphics[width=\textwidth]{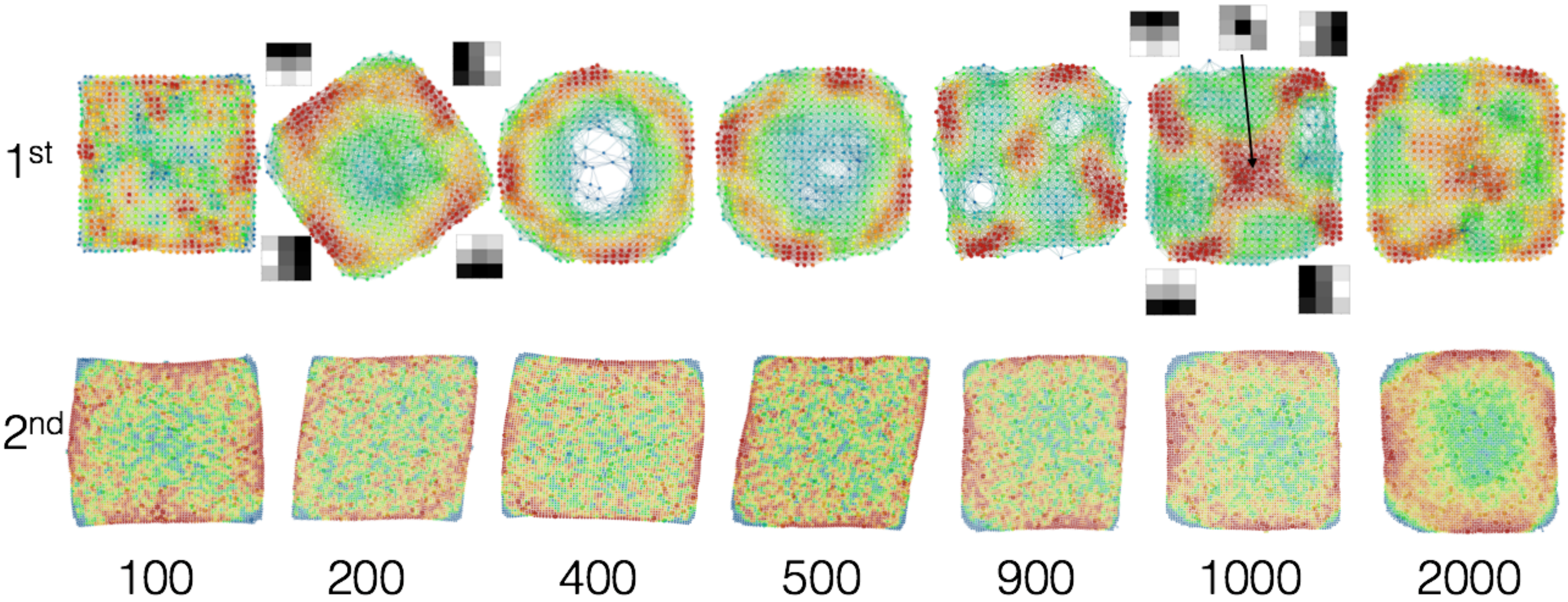}
\caption{CIFAR-10 learning}\label{learning}
\end{center}
\end{minipage} 
\end{figure}}

Finally, we examined a well known pretrained neural network, VGG16, trained on Imagenet, a large image data base \cite{vgg16},\cite{imagenet}.  This neural net has 13 convolutional layers, and so permits us to study seriously the ``responsibilities" of the various layers. 
{\begin{figure}[!htp]
\centering
\begin{minipage}[t]{.7\textwidth} 
\begin{center}
\includegraphics[width=\textwidth]{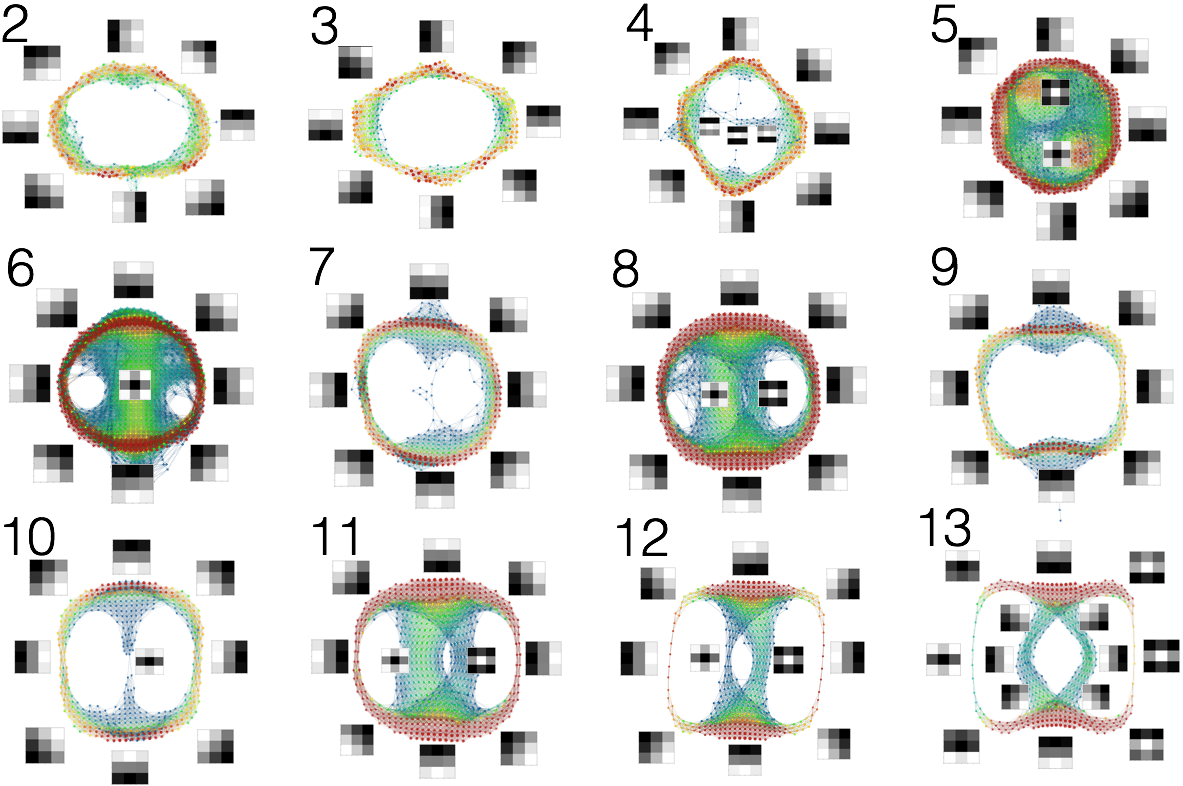}
\caption{VGG16}\label{VGG16}
\end{center}
\end{minipage} 
\end{figure}} 
Mapper models of the sets of weight vectors for layers 2-13 are shown in Figure \ref{VGG16}.  In this case, the neural net has sufficiently many grids in each layer to construct a useful data set from this network alone.  Observe that the first two layers give exactly a primary circle, and that after that more complex things appear.  Secondary circle patches occur in layer 4, and in higher layers we see different phenomena occurring, including the bullseye we saw in CIFAR-10, as well as crossings of lines.  One interesting line of research would be to assemble all of these different phenomena into a single space, including the Klein bottle.  The advantage of doing this is that it will permit feature generation in terms of functions on the space, such as was done in \cite{texture}, or improved compression algorithms as in \cite{maleki}.  For now, the outcome demonstrates with precision how the higher layers encode higher layers of abstraction in images, as occurs in the mammalian visual pathway.

\section{Feature Geometries and Architectures}
\subsection{Generalities}
Since CNN's have demonstrated a great deal of success on data sets of images, the idea of trying to generalize it suggests itself.  To perform the generalization, one must identify what properties of image data sets are being used, and how.  There are two key properties.  

\begin{itemize}
\item{{\bf Locality:} The features in image data set (i.e.pixels) are equipped with a geometry, i.e. that of a rectangular grid.  That grid is critical in restricting the connections in the corresponding feed-forward structure, and that restriction can be formulated very simply in terms of the $L^{\infty}$  distance function on the grid, as we have seen in our constructions of CNN's in Section \ref{findings}. This observation suggests  that one can use other metric spaces to restrict the connections in architectures based on these metric spaces.   We note that the grid geometry can be regarded as a discretization of the geometry of the plane, or of a square in the plane.   }
\item{{\bf Homogeneity:} The convolutional neural net is equipped not only with a connection structure, but a choice of  {\em convolutional structure} (as in Definition \ref{defcon}), which creates its own restrictions on the features created in the neural net.  Because it requires that weight vectors associated with one point in the grid be identical with those constructed at other points, the convolutional property should be interpreted as a kind of homogeneity.  In addition to putting drastic limitations on the features being created in the neural net, this restriction encodes a property of image data sets that we regards as desirable, namely that the same object occurring in different parts of an image should be detected in an identical fashion. }
\end{itemize} 

What we would like to do is to describe how the two properties above can be used to construct neural nets in an analogous fashion, to improve performance on image data sets and to generalize the ideas to more general data sets.  In order to have a notion of locality, we will need to understand data sets in terms of the geometry of their sets of features.  We identify at least three methods in which feature sets can obtain a geometry.  

\begin{enumerate} 
\item{{\bf A priori geometries:} The prime example here is the image situation, where the grid geometry is explicitly constructed in the construction of the data.  The continuous version of this geometry is that of the plane.    Other examples would include time series, where the a priori continuous geometry is  the line, or periodic time series, where the geometry is that of the circle.  The geometries for the building of the neural net would be discretizations of these geometries, obtained by selecting discrete subsets, often in a regular way. }
\item{{\bf Geometries obtained from data analysis:} The data analysis performed in \cite{klein} or \cite{gabrielsson} reveals that the frequently occurring local patches in images concentrate around a primary circle, and that these patches are well modeled by particular functions which can be algebraically defined.  We will show below that this fact permits the construction of a set of features for images which admit a circular geometry.    One could also construct a Klein bottle based set of features and a corresponding Klein bottle based geometry on that set. }
\item{{\bf Purely data drive geometries:} In many situations one does not want  to perform a detailed modeling procedure for the set of features, but nevertheless wants to use feature geometries to restrict connections in neural nets which are designed to learn a function based on the features.  In this case, one can use the Mapper methodology \cite{mapper} to obtain  discretized versions of geometries on the feature space, well suited to the construction of neural nets.  }
\end{enumerate} 

Section \ref{images} can be regarded as a discussion of one case where an a priori geometry is available, so we will not discuss it further.  Instead, we will  give examples of data analytically obtained geometries and purely data driven constructions. 

\subsection{Data-analytically Defined Geometries} \label{dataanalytic}

 We  first consider the data analytic work that was done in \cite{klein} and \cite{gabrielsson}.  We find that the frequently occurring patches are approximable by discretizations of linear intensity functions onto a $3 \times 3$ grid.  To be specific, we regard the pixels in a $3 \times 3 $ patch to be embedded in the square $I^2 = [-1,1] \times [-1,1]$, as the subset ${\cal{L}} =  \{ -1,0,1 \} \times \{ -1, 0 , 1 \}$.  The discretization operation can be considered as the restriction of a function on $I ^2$ to ${\cal L}$.  We consider the set of linear functions in two variables given by the formulae
$$  f_{\theta} (x,y) = x\cos (\theta ) + y \sin (\theta) 
$$
The set of functions is parametrized by the circle valued parameter $\theta$.  For each $f_{\theta}$, we can construct a function on an image as follows.  Let $(m,n) \in \bb{Z}^2$ denote a particular pixel in the grid defining an image data set ${\cal D} $ consisting of images $p$,  with  $p(m,n)$ denoting  the gray scale value of an image within the data set.  Given an angle $\theta$, we now define a function $q_{m,n,\theta}(p) $ on ${\cal D}$ by the formula 
$$  q_{m,n,\theta}(p)  = \sum _{(i,j) \in {\cal L}} p(m+i,n+j)\cdot f_{\theta}(i,j)
$$
In this case the continuous geometry associated to the feature space for these images is $\bb{R}^2 \times S^1$.  The discretization will be choosing a rectangular lattice $L$ for $\bb{R}^2$ in the usual way, and by choosing the set $\mu _n$  of $n$-th roots of unity for the circular factor.  So the discretized form is $\bb{Z}^2 \times \mu _n$. This set is a metric space in its own right, and we can use the metric correspondences defined in Example \ref{metriccorr} to construct generators and neural nets based on this geometry.  
\begin{Remark}{\em There are similar synthetic models with a Klein bottle $K$ replacing $S^1$.  There are natural choices for discretizations of $K$ as well.  }
\end{Remark}

We have demonstrated that there are methods of imposing locality on new features that have been constructed based on the data analysis of image patches and of weight vectors in convolutional neural nets.  For this construction, there are also convolutional structures as defined in Definition \ref{defcon}.
In fact, they are Cayley structures in the sense of Example \ref{convdef}, as we can readily see from the observation that the metric space $\bb{Z}^2 \times \mu _n$ is equipped with a free and transitive action by the group $\bb{Z}^2 \times \bb{Z}/n\bb{Z}$, and this group action determines a Cayley convolutional structure.   This gives a number of possibilities for the construction of new feed-forward systems  with feature geometries taken in to account.  To see how these might look, let's consider the feed-forward system $F$ described in (\ref{elementary}) above.   $F$ is broken into a product $F = F^s \times F^c$, where $F^c$ is a complete generator, and the structural factor $F^s$ is given by 
$$  \bb{Z}^2  \xrightarrow{\footnotesize\makebox[1.2cm]{ ${\cal C}_d(1)$}} \bb{Z}^2 
  \xrightarrow{\footnotesize\makebox[1.8cm]{ $\pi^2(0,1,2)$}} \bb{Z}^2   \xrightarrow{\footnotesize\makebox[.8cm]{ ${\cal C}^c$}} X(10)
$$
The idea will be to construct new structural factors by taking products with generators involving only $\mu _n$ for various $n$'s. We'll call these generators {\em angular factors}.   The simplest one is of the form 
$$  \mu_n   \xrightarrow{\footnotesize\makebox[.5cm]{${\cal C}^c$}} X(1)  \xrightarrow{\footnotesize\makebox[.5cm]{${\cal C}^c$}}  X(1) \xrightarrow{\footnotesize\makebox[.5cm]{${\cal C}^c$}}  X(1) 
$$
Here $X(1)$ denotes a one element set.  The corresponding structural factor including the grids would then be 
$$ \bb{Z}^2 \times \mu_n   \xrightarrow{\footnotesize\makebox[1.8cm]{${\cal C}_d(1) \times {\cal C}^c$}} \bb{Z}^2 \times X(1) \simeq \bb{Z}^2  \xrightarrow{\footnotesize\makebox[1.8cm]{ $\pi^2(0,1,2)$}}  \bb{Z}^2 \xrightarrow{\footnotesize\makebox[.8cm]{${\cal C}^c$}} X(10) \times  X(1) \simeq X(10)
$$
The effect of this modification is simply to use the newly constructed features directly in the computation.  It permits the algorithm to use them directly rather than having to ``learn" them.  Another angular factor is 
$$  \mu_n   \xrightarrow{\footnotesize\makebox[.5cm]{${\cal C}^c$}}\mu _n   \xrightarrow{\footnotesize\makebox[.5cm]{${\cal C}^c$}}  X(1) \xrightarrow{\footnotesize\makebox[.5cm]{${\cal C}^c$}}  X(1) 
$$
Forming the product of this angular factor with $F^s$ and ultimately $F^c$ as well produces a feed-forward structure which creates new angular factors in layer $1$.  The corresponding neural networks would be able to learn angle dependent functions from earlier angular functions. Yet another angular factor would be the following. 
$$  \mu_n   \xrightarrow{\footnotesize\makebox[.5cm]{${\cal C}_d(\xi)$}}\mu _n   \xrightarrow{\footnotesize\makebox[.5cm]{${\cal C}^c$}}  X(1) \xrightarrow{\footnotesize\makebox[.5cm]{${\cal C}^c$}}  X(1) 
$$
where $\xi $ is the distance from $(1,0)$ to the primitive root of unity $\zeta _n = (\cos(\frac{2\pi}{n}) , \sin(\frac{2 \pi}{n}))$.  Adding this angular factor to $F^s$ creates new angular features  in layer $1$, allows these angular features to learn from angular and raw features, and further restricts that learning so that a given angular feature would only depend on raw values and angular features in the input that are near to the given feature in the metric on $\mu _n$.  This is the angular analogue to the idea that a convolutional neural net permits a feature in a convolutional layer to depend only on features in the preceding layer that are spatially close to the given feature, in this case in the a priori geometry on pixel space.  

There is also an analogue for $\mu _n$  to the pooling correspondences $\pi(m,n,N)$ defined in Section \ref{images}.  They are correspondences from $\pi _{m,n}:\mu _{mn} \longrightarrow \mu _{n}$, and they are defined by 
$$  \pi _{m,n}(\zeta _{mn} ^x) =  \zeta _n^{\lfloor \frac{x}{m} \rfloor}
$$
It is easy to verify that this is well-defined.  We have only created analogues to the correspondences $\pi (0,n-1,n)$ from Section \ref{images}, but analogues for other values of $m,n$,  and $N$  exist as well.  We could now construct a new angular factor 
$$  \mu_{mn}   \xrightarrow{\footnotesize\makebox[.8cm]{${\cal C}_d(\xi)$}}\mu _{mn}   \xrightarrow{\footnotesize\makebox[.8cm]{$ \pi _{m,n}$}} \mu _n \xrightarrow{\footnotesize\makebox[.5cm]{${\cal C}^c$}}  X(1) 
$$
which would incorporate pooling in the angular directions as well.  Each of these constructions have analogues for the case of the Klein bottle geometries.  

We have some preliminary experiments involving the simplest versions of these geometries.  We have used them to study MNIST, as well as the SVHN data set \cite{svhn}.  SVHN is another data set of images of digits, collected by taking photographs of house numbers on mailboxes and on houses. For these studies, we have simply modified the feed-forward systems by constructing the product of the existing structural factors described in (\ref{mnistnetwork}) and (\ref{cifarnetwork}) with an additional structural factor of the form 
\begin{equation}\label{angularfactor}( \mu _{16})_+ \xrightarrow{\footnotesize\makebox[.5cm]{${\cal C}^c$}} X(1) \xrightarrow{\footnotesize\makebox[.5cm]{${\cal C}^c$}} X(1)
\xrightarrow{\footnotesize\makebox[.5cm]{${\cal C}^c$}} X(1) 
\xrightarrow{\footnotesize\makebox[.5cm]{${\cal C}^c$}} X(1) 
\xrightarrow{\footnotesize\makebox[.5cm]{${\cal C}^c$}} X(1)\xrightarrow{\footnotesize\makebox[.5cm]{${\cal C}^c$}} X(1)
\end{equation}
where $(\mu_{16})_+ $ plus denotes $\mu _{16}$ with a disjoint point added.  This additional point is there so that we include the original ``raw" pixel features.  
This amounts to including the ``angular" coordinates described above as part of the input data, and using it to inform the higher level computations.  We have two results, one in the direction of speeding up the learning process and the other concerning the generalization from a network trained on MNIST to SVHN.  

\begin{itemize}
\item{We found substantial improvement in the training time for both MNIST and SVHN when using the additional angular features.  A factor of $2$ speed up was realized for MNIST, and a factor of $3.5$ for SVHN.  MNIST is a much cleaner and therefore easier data set, and we suspect that the speed up will in general be larger for more complex data sets.   }
\item{We also examined the degree to which a network trained on one data set (MNIST) can achieve good results on another data set (SVHN).  Using the standard convolutional network for images, we found that a model trained on MNIST applied to SVHN achieved roughly $10\%$ accuracy. Since there are 10 distinct outcomes, this is essentially the same as selecting a classification at random. However, when we built the corresponding model using the additional factor (\ref{angularfactor}) above, we found that the accuracy improved to $22\%$.  Of course, one wants much higher accuracy, but what this finding demonstrates is that this generalization problem can be substantially improved using these methods.   }
\end{itemize} 
In these examples, we have only used the simplest versions of the constructions we have discussed in Section \ref{neuralnets}.  The possibilities that we envision going forward include taking products with structural factors of the form 
\begin{equation} \label{betterstructural}(\mu_{4n})_+ \xrightarrow{\footnotesize\makebox[.9cm]{${\cal C}_d(\xi_{4n})_+$} }(\mu_{4n})_+ \xrightarrow{\footnotesize\makebox[.8cm]{$(\pi _{4n,2n})_+$}}(\mu_{2n})_+
\xrightarrow{\footnotesize\makebox[.7cm]{${\cal C}_d(\xi_{2n})_+$}}(\mu_{2n})_+
\xrightarrow{\footnotesize\makebox[.8cm]{$(\pi_{2n,n})_+ $}} (\mu_{n})_+ 
\xrightarrow{\footnotesize\makebox[.5cm]{${\cal C}^c $}}X(1) \xrightarrow{\footnotesize\makebox[.5cm]
{${\cal C}^c $}}X(1) 
\end{equation}
The correspondences ${\cal C}_d(\xi_{2^i n})_+$ and $(\pi _{2^in,2^{i-1}n})_+$ for $i = 0,1,2$ in this feed-forward system are straightforward generalizations of ${\cal C}_d(\xi _{2^in})$ and $\pi _{2^in,2^{i-1}n}$ to the situation where the disjoint base point $+$ has been added.  $({\cal C}_d(\xi _{n}))_+$ is obtained by constructing a metric on $(\mu _n)_+$ for which the distance from the point $+$ to each of the elements of $\mu _n$, as well as all the distances between adjacent roots of unity, are all equal to $\xi _n$.  It is not hard to see that this can be done.  $(\pi _{2n,n})_+$ is the functional correspondence which is equal to $\pi _{2n,n}$ on $\mu _n$ and which carries the point $+$ to $+$. The effect of this construction is that it would include angular features at the higher layers, and that it would restrict the angular  features that are constructed to include only those which involve nearby angular features in the preceding layers.

\subsection{Purely Data Driven Geometries}

Suppose that we are given a data set defined by a data matrix $D$, with the rows corresponding to the data points and the columns corresponding to the features, but that we have no theory for the features  analogous to the one described in \cite{klein}.  What we generally have, though, are metrics on the set of features.  If the matrix entries are continuous, one can use Euclidean distance of the features viewed as column vectors.  There are variants, such as mean centered and/or variance normalized versions, correlation distance, angle distance, etc.  If the entries of the matrix are binary, then Hamming distance is an option.  In general, it is most often possible to define, in natural ways, metrics on the set of columns.  This means that the feature set is a metric space, and therefore that we already have the possibility of carrying out part of the process used on image data sets, namely the construction of the correspondences ${\cal C}_d(r): X \rightarrow X$, where $X$ denotes the feature set. We refer to the column space equipped with a metric as the {\em feature space}.  These can be used to create a counterpart for the initial  convolutional layers in the feed-forward system, but it does not give a counterpart to the pooling correspondences.  The pooling correspondences are important because they allow one to study features that are more broadly distributed in the geometry of the feature space.  To construct deeper networks, one may also need an analogue for higher level convolutional layers.  There is an approach using the Mapper methodology introduced in \cite{mapper} that will directly construct a counterpart to pooling methodology.  

We recall that the output of Mapper, applied to a finite metric space $X$, is a graph $\Gamma$, with an assignment to each vertex $v$ of $\Gamma$ a subset $X_v$ of $X$, having the following two properties:
\begin{enumerate}
\item{Every point $x  \in X$ is contained in $X_v$ for some vertex $v$ of $\Gamma$. }
\item{Two vertices $v$ and $w$ of $\Gamma$ are connected by an edge if and only if $X_v \cap X_w \neq \emptyset$.   }
\end{enumerate}
We observe that this means that if we have two Mapper models  $(\Gamma, \{X_v \}_{v \in V(\Gamma ) })$ and   $(\Gamma^{\pr}, \{X^{\pr}_{w}\}_{w\in V(\Gamma ^{\pr} ) })$ on the same metric space $X$, then there is a well-defined correspondence $${\cal C}(\Gamma, \Gamma ^{\pr}) : V(\Gamma) \rightarrow V(\Gamma ^{\pr})$$
defined by the property that for $(v,w) \in V(\Gamma) \times V(\Gamma ^{\pr})$, $(v,w) \in {\cal C}(\Gamma, \Gamma ^{\pr}) : V(\Gamma) \rightarrow V(\Gamma ^{\pr})$  if and only if $X_v \cap X^{\pr}_w \neq \emptyset$.   

These properties allows us to construct two specific correspondences.  Given a metric space $X$ and a Mapper model $(\Gamma, \{X_v \}_{v \in V(\Gamma ) })$ for the feature space of a data matrix, we have the {\em augmentation correspondence} $\varepsilon: X \rightarrow V(\Gamma )$, defined by $(x,v) \in X \times V(\Gamma)$ if and only if $x \in X_v$.   We also have the correspondence ${\cal C}(\Gamma,\Gamma) : V(\Gamma) \rightarrow V(\Gamma)$. 
\begin{Remark} {\em The correspondence ${\cal C}(\Gamma, \Gamma)$ is simply the graph correspondence ${\cal C}_{\Gamma}$ defined in Example \ref{graphcorr}.   }
\end{Remark}
To define analogues to pooling correspondences,  we need a bit more detail on the Mapper construction.  It begins with one or more projections $f: X \rightarrow \bb{R}$, which we call {\em filters}.  Typically there are only a small number of $f$'s, perhaps 1,2, or 3, and we denote the collection of filters by $\{ f_{\alpha} \}_{\alpha \in A}$, where it is understood that $\#A$ is small.    We now construct a family of  open coverings of the real line. 

\begin{Definition}  Given a pair  $( l ,s)$ of real numbers with $l > s$, we define the covering ${\cal U}(l,s)$ to consist of all intervals of the form $(ks-\frac{l}{2},ks  + \frac{l}{2})$, $k \in \mathbb{Z}$.  The condition $l>s$ guarantees that the family is a covering.  Given a pair $( l, s)$, we defined the {\em double } of $( l ,s)$ to be the pair $( 2l, 2s)$.  ${\cal U}( 2l,2s )$ covers $\bb{R}$ with intervals of double the length of the intervals comprising  ${\cal U}(l,s)$.  We refer to $l$ as the {\em length}  and $s$ as the {\em stride}.   
\end{Definition} 

Let $n$ denote the cardinality of $A$, and equip $A$ with a total ordering, so $A = \{ \alpha _1, \ldots , \alpha _n \}$.  Let $F: X \rightarrow \bb{R}^n$ denote the product $f_{\alpha _1} \times \cdots \times f_{\alpha _n}$.  For each filter $f_{\alpha}$, we choose a pair  $( l _{\alpha}, s _{\alpha})$.   For each $\alpha \in A$, we let ${\cal U}_{\alpha} = {\cal U}( l _{\alpha}, s _{\alpha})$, and let ${\cal U}_{\alpha} = \{ I_{\beta _{\alpha} }^{\alpha} \}_{\beta _{\alpha}  \in B_{\alpha}}$, where $B_{\alpha}$ is an indexing set for the intervals in ${\cal U}_{\alpha}$.   We now construct the product covering ${\cal U}_{\alpha _1} \times {\cal U}_{\alpha _2} \times \cdots \times {\cal U}_{\alpha _n}$  of $\bb{R}^n$, which consists of sets of the form $I^{\alpha _1}_{\beta _{\alpha _1}} \times \cdots \times I^{\alpha _n}_{\beta _{\alpha _n}} $, for all choices of $n$-tuples $(\beta _{\alpha _1}, \ldots , \beta _{\alpha _n})$ in $B _{\alpha _1 }\times \cdots \times B_{\alpha _n}$.  We denote this covering by ${\cal V} = \{ V_j \}_{j \in J}$, where $J$ is the indexing set.  We now create overlapping subsets (bins) of the form $F^{-1}(V_j)$.  These sets form a covering of $X$.  The algorithm defined in \cite{mapper} next proceeds by clustering (using a predefined clustering algorithm) each of the bins, creating a partition of each bin.  The vertex set of the Mapper  model  $\Gamma$ of $X$ consists of one element for each block of each partition of each bin, and we declare that two vertices are connected by an edge if and only if the corresponding blocks overlap.  Note that the blocks can overlap because the bins overlap.  It is clear from the construction that this construction has the two properties ascribed to it above.  
Note that the construction described above depended only on the choices $(l_{\alpha}, s_{\alpha})$.  The result of applying the above construction to the choices $(2l_{\alpha}, 2s_{\alpha})$ will be referred to as the {\em double} of $\Gamma$, and we will denote it by $\Gamma (1)$, where it is understood that $\Gamma = \Gamma (0)$.  We can iterate this process to obtain a sequence of Mapper models 
$ \Gamma (0), \Gamma (1), \ldots , \Gamma(r)
$, where $\Gamma (i+1)$ should be viewed as a ``coarsening" of $\Gamma (i) $  or ``lower resolution model" than $\Gamma(i)$.  Just as we use pooling correspondences to pass from a higher resolution image to a lower resolution image, so we can now use the correspondences ${\cal C}(\Gamma (i), \Gamma(i+1))$ as methods from passing to high resolution to lower resolution versions of the feature space for an arbitrary data matrix.  

We show how this will work by constructing an analogue of the depth $6$ generator constructed for MNIST in (\ref{mnistnetwork}) above.  We suppose that we have selected a metric on the column space of our data matrix, and further that we have built a Mapper model $\Gamma  =  \Gamma (0)$, together with the doublings $\Gamma (1) $ and $\Gamma (2)$.  Further, we suppose we are trying to solve an $N$-outcome classification problem, where $N$ was $10$ in the actual MNIST case.  As in the MNIST case, the generator  will decompose as a product of a complete generator $F^c$ and a structural generator $F^s$.  The complete generator can be chosen arbitrarily.  The analogue to the structural generator in (\ref{mnistnetwork}) is given by the following. 
$$X   \xrightarrow{\makebox[.4cm]{$\varepsilon$}}\Gamma (0)   \xrightarrow{\makebox[1.8cm]{\footnotesize${\cal C}(\Gamma (0),\Gamma (1))$}}\Gamma (1)  \xrightarrow{\makebox[1.8cm]{\footnotesize${\cal C}(\Gamma (1),\Gamma (1))$}} \Gamma(1)  \xrightarrow{\makebox[1.8cm]{\footnotesize${\cal C}(\Gamma (1),\Gamma (2))$}} \Gamma (2)  \stackrel{{\cal C}^c}{\longrightarrow }X(1) \stackrel{{\cal C}^c}{\longrightarrow} X(N)
$$
Unlike the data analytically driven neural nets, this construction has not yet been done but is in development.  

Finally, we point out that one need not adhere rigidly to the doubling strategy described above.  Choosing any families of coverings that are increasing, in the sense that $l$ and $s$ are both increasing, also can give families of correspondences that can act as replacements for pooling correspondences.

 \end{document}